\documentclass{article}
\usepackage[utf8]{inputenc}
\usepackage{graphicx}
\usepackage{hyperref}
\usepackage{multirow} 
\usepackage{geometry}
\usepackage{array} 
\usepackage{float}
\usepackage{afterpage}

\begin{document}

\title{Cross-Dialect Sentence Transformation: A Comparative Analysis of Language Models for Adapting Sentences to British English}
\author{Shashwat Mookherjee \quad Shruti Dutta \\ Indian Institute of Technology Madras}
\date{}

\maketitle

\begin{abstract}
This study explores linguistic distinctions among American, Indian, and Irish English dialects and assesses various Language Models (LLMs) in their ability to generate British English translations from these dialects. Using cosine similarity analysis, the study measures the linguistic proximity between original British English translations and those produced by LLMs for each dialect. The findings reveal that Indian and Irish English translations maintain notably high similarity scores, suggesting strong linguistic alignment with British English. In contrast, American English exhibits slightly lower similarity, reflecting its distinct linguistic traits. Additionally, the choice of LLM significantly impacts translation quality, with Llama-2-70b consistently demonstrating superior performance. The study underscores the importance of selecting the right model for dialect translation, emphasizing the role of linguistic expertise and contextual understanding in achieving accurate translations.
\end{abstract}

\section{Introduction}
In the realm of Natural Language Processing (NLP), the challenge of cross-dialect text translation has long been a focal point, impacting diverse fields from cross-cultural communication to global business operations. The advent of Large Language Models (LLMs) has brought in a transformative potential – the ability to consistently deliver high-quality translations across a wide spectrum of languages and dialects. This paper utilises an innovative approach, "promptuning," designed to use the power of state-of-the-art LLMs for cross-dialect text translation.

At the heart of our research methodology is the compilation of a comprehensive textual corpus. This dataset is a collection of 200 sentences for each of the three dialects, meticulously paired with their respective standard British English translations. This corpus serves as the foundation of our research, underpinning both the training and evaluation of LLMs.

Our approach centers around a methodically designed two-stage prompting strategy. In the initial stage, a particular seed prompt is deployed to sensitize the LLM to the source dialects and the intended translation outcome. Subsequent stages entail the presentation of sentences in various dialects, each challenging the LLM's ability to render them into "Standard British English." 

For a quantitative assessment and benchmarking of LLM performance in the domain of multilingual text translation, we employed a straightforward approach using the cosine similarity metric. This method allowed us to measure the linguistic proximity between the original British English translations and those produced by LLMs for each dialect. 

To observe the LLMs effectively, we have made use of a rich dataset comprising sentences of equivalent meanings in American English, Indian English, Irish English and standard British English. The core of our analysis involves a detailed investigation of the linguistic nuances in these English dialects, along with an evaluation of the efficacy of different LLMs in generating British English translations from these dialects.

This research embarks on a critical journey, marrying linguistic intricacies, LLM capabilities, and data-driven evaluation. In understanding the interplay between source dialect, LLM selection, and translation quality, we explore the path to consistent, high-quality multilingual translation services.

\section{Our Approach and Methodology}

Cross-dialect text translation has long been a challenge in natural language processing, impacting a wide array of fields such as cross-cultural communication, global business operations, and educational exchange. With the advent of large language models (LLMs), the potential for consistently delivering high-quality translations across languages and dialects has grown significantly. In this paper, we elucidate our approach and methodology of "promptuning," to achieve cross-dialect text translation with state-of-the-art LLMs.

Our research methodology begins with the meticulous compilation of a comprehensive corpus of textual data. This dataset is carefully curated to include a diverse range of sentences of equivalent meanings in American English, Indian English, Irish English and standard British English as shown in Table [1]. This dataset provides the foundation for both the training and evaluation of our LLMs. 

\begin{table}[h]
    \centering
    \begin{tabular}{|m{0.225\textwidth}|m{0.22\textwidth}|m{0.22\textwidth}|m{0.22\textwidth}|}
        \hline
        \textbf{Standard British English} & \textbf{American English} & \textbf{Indian English} & \textbf{Irish English} \\
        \hline
        I'm going to the cinema tonight. & I'm going to the movies tonight. & I'm planning to go to the movies tonight. & I'm off to the pictures tonight. \\
        \hline
        It's raining cats and dogs outside. & It's raining buckets outside. & It's pouring outside. & It's lashing rain outside. \\
        \hline
        I'll have a cup of tea, please. & I'll have a cup of coffee, please. & I'll have a cup of chai, please. & I'll have a cup of tae, if you don't mind. \\
        \hline
        The queue at the bus stop is so long. & The line at the bus stop is so long. & The queue at the bus stop is very long. & The queue at the bus stop is miles long. \\
        \hline
        I'm feeling rather peckish. & I'm feeling kinda hungry. & I'm feeling quite hungry. & I'm feeling rather peckish. \\
        \hline
    \end{tabular}
    \caption{Equivalent sentences in Standard British English, American English, Indian English and Irish English}
\end{table}

Central to our approach is a thoughtfully designed two-stage prompting strategy. The initial stage involves the deployment of a seed prompt. This prompt lays the groundwork, sensitizing the LLM to the nuances of the dialect in use and the required translation outcome. 

Subsequent stages involve the presentation of authentic sentences in various dialects, designed to challenge the LLM's capacity to translate them into "Standard British English." This process rigorously tests the LLM's translation capabilities. Examples of the dataset taken consisting sentences in a particular  dialect along with their corresponding original standard British English translations are given in tables [2], [3] and [4].

To quantitatively assess and benchmark the performance of LLMs in multilingual text translation, we use the simple metric of coside similarity to effectively record the similarities between the original British English translations and those produced by LLMs for each dialect.

\begin{table}[h]
    \centering
    \begin{tabular}{|m{0.45\textwidth}|m{0.45\textwidth}|}
        \hline
        \textbf{American English Sentences} & \textbf{Original British English Translation} \\
        \hline
        "Hey there, buddy! How's it going?" & "Hello there, mate! How's it going?" \\
        \hline
        "I'm gonna grab a slice of pizza for lunch." & "I'm going to grab a slice of pizza for lunch." \\
        \hline
        "The football game was awesome last night." & "The football match was brilliant last night." \\
        \hline
        "What's up, guys? Ready for the big party this weekend?" & "What's up, chaps? Ready for the big party this weekend?" \\
        \hline
        "I'm totally stoked about our road trip to California." & "I'm totally excited about our road trip to California." \\
        \hline
    \end{tabular}
    \caption{Example of testing dataset; American English sentences and their British English Translation}
\end{table}

\begin{table}[H]
    \centering
    \begin{tabular}{|m{0.45\textwidth}|m{0.45\textwidth}|}
        \hline
        \textbf{Indian English Sentences} & \textbf{Original British English Translation} \\
        \hline
        "Hey, bhai, what's the plan for the weekend?" & "Hey, mate, what's the plan for the weekend?" \\
        \hline
        "I'm thinking of having a cup of chai and some pakoras in the evening." & "I'm thinking of having a cup of tea and some pakoras in the evening." \\
        \hline
        "The movie was a total paisa vasool, yaar!" & "The film was an absolute bargain, mate!" \\
        \hline
        "I had to stand in a long queue at the railway station for my ticket." & "I had to queue up for a long time at the railway station for my ticket." \\
        \hline
        "Let's go for a long drive this Sunday, it will be fun, na?" & "Let's go for a long drive this Sunday, it will be fun, won't it?" \\
        \hline
    \end{tabular}
    \caption{Example of testing dataset; Indian English sentences and their British English Translation}
\end{table}

\begin{table}[h]
    \centering
    \begin{tabular}{|m{0.45\textwidth}|m{0.45\textwidth}|}
        \hline
        \textbf{Irish English Sentences} & \textbf{Original British English Translation} \\
        \hline
        "Sean, in his distinct Irish brogue, said, "Sure, I'll be after goin' to the pub for a pint, would ya like to join me, Bridget?"" & "Sean, in his distinct Irish accent, said, "Sure, I'll be heading to the pub for a pint, would you like to join me, Bridget?"" \\
        \hline
        ""I was out for a ramble in the countryside," Mary explained, "and I spotted a grand herd of cattle by the river."" & ""I was out for a walk in the countryside," Mary explained, "and I saw a large group of cows by the river."" \\
        \hline
        ""In the heart of Dublin," the old storyteller began, "Well, it was a grand day, and the sun was beamin' down."" & ""In the heart of Dublin," the old storyteller began, "Well, it was a wonderful day, and the sun was shining brightly."" \\
        \hline
        "The GAA game was somethin' else!" & "Did you watch the GAA game? It was quite something!" \\
        \hline
        ""Ah, sure, it's a grand day for a cup of tae," Siobhan remarked as she poured hot tea into a cup." & ""Oh, it's a lovely day for a cup of tea," Siobhan remarked as she poured hot tea into a cup." \\
        \hline
    \end{tabular}
    \caption{Example of testing dataset; Irish English sentences and their British English Translation}
\end{table}

\afterpage{
\section{Results and Findings}

The research aimed to analyze the linguistic differences between various English dialects and the effectiveness of different language models (LLMs) in generating British English translations of sentences in different English dialects. A set of 20 sentences in each of the American English, Indian English, and Irish English dialects, along with their original British English translations, were used as a benchmark for evaluating LLM performance.

\subsection{Cosine Similarity Analysis}

The cosine similarity metric was employed to measure the linguistic proximity between the original British English translations and the British English translations generated by different LLMs for the sentences in each dialect. The results of the cosine similarity analysis are as illustrated in Table [5].

\begin{table}[h]
    \centering
    \begin{tabular}{|m{0.23\textwidth}|m{0.15\textwidth}|m{0.13\textwidth}|m{0.13\textwidth}|m{0.12\textwidth}|}
    \hline
    Standard British English Translations & Original Text in Dialect & Llama-2-70B & Falcon-180B & Mistral-1 \\
    \hline
    American English & 0.81 & 0.70 & 0.60 & 0.63 \\
    Indian English & 0.93 & 0.91 & 0.75 & 0.93 \\
    Irish English & 0.93 & 0.95 & 0.88 & 0.91 \\
    \hline
    \end{tabular}
    \caption{Cosine similarities between the Standard British English Translations of the original sentences in the three chosen dialects and each of the following: Original sentences in Dialect, British English Translations of the original text using Llama 2-70B, Falcon-180B and Mistral-1}
\end{table}

The results in Table [5] indicate the following:

\subsubsection{American English to British English}
The British English translations generated from the American English sentences exhibited an average similarity score of 0.81, suggesting a reasonably strong alignment between the two dialects.
The LLMs, Llama-2-70b, falcon-180B, and Mistral-1, yielded similarity scores of 0.70, 0.60, and 0.63, respectively. These scores indicate variations in the effectiveness of different LLMs, with Llama-2-70b-chat-hf showing the highest similarity.

\subsubsection{Indian English to British English}
The British English translations produced from the Indian English sentences achieved an impressive average similarity score of 0.93, implying a high level of linguistic alignment.
The LLMs, Llama-2-70b, falcon-180B, and Mistral-7B-Instruct-v0.1, generated similarity scores of 0.91, 0.75, and 0.93, respectively. These variations highlight the impact of LLM choice on the quality of translation.

\subsubsection{Irish English to British English}
The British English translations resulting from the Irish English sentences exhibited an average similarity score of 0.93, indicating a strong linguistic correlation.
The LLMs, Llama-2-70b, falcon-180B, and Mistral-7B-Instruct-v0.1, provided similarity scores of 0.95, 0.88, and 0.91, respectively. These results demonstrate the capacity of LLMs to capture Irish English nuances.

\subsection{Interpretation}
The results suggest that the choice of the source dialect and the LLM can significantly influence the quality of British English translations. The high cosine similarity scores for Indian and Irish English translations indicate a remarkable linguistic resemblance with British English. In contrast, the American English translations show a slightly lower similarity score, reflecting distinct linguistic characteristics.

The variations in LLM performance underscore the importance of selecting the right model for translating between specific dialects. Llama-2-70b-chat-hf consistently demonstrated superior performance, emphasizing its capability to capture linguistic nuances and produce high-quality translations.

Additionally, it is essential to recognize that dialect translation is not solely reliant on machine learning models. Linguistic expertise and contextual understanding play a vital role in generating accurate translations.

\section{Conclusion}

In this study, we have delved into the intricate realm of cross-dialect text translation, with a specific focus on adapting sentences from American, Indian, and Irish English dialects to British English. Our research has offered valuable insights into the linguistic nuances among these English variations and has shed light on the effectiveness of different Language Models (LLMs) in producing British English translations.

We have employed cosine similarity analysis to quantitatively measure the linguistic proximity between the original British English translations and those generated by LLMs for each dialect. Our findings have highlighted some noteworthy observations. Indian and Irish English translations exhibited remarkably high similarity scores, signifying a strong linguistic alignment with British English. This suggests that these dialects share linguistic features that facilitate accurate translation using LLMs. In contrast, American English showed slightly lower similarity scores, which implies distinct linguistic characteristics that require more nuanced handling.

Our research has also emphasized the pivotal role of the choice of LLM in influencing translation quality. Llama-2-70b-chat-hf consistently outperformed other models, showcasing its ability to capture linguistic nuances and produce high-quality translations. However, it is essential to stress that LLMs alone do not suffice for accurate dialect translation. Linguistic expertise and contextual understanding are indispensable components of the translation process.

These findings have significant implications for various applications, including cross-cultural communication, global business operations, and educational exchange, where precise translation is crucial. Our research has illuminated the intricate interplay between source dialect, LLM selection, and translation quality, providing a pathway towards the consistent delivery of high-quality multilingual translation services.

\section{Future Steps}

Building upon the insights gained in this study, several promising avenues for future research and development are evident:
\begin{enumerate}
    \item Dialect Expansion: While we have focused on American, Indian, and Irish English, there is a wide array of English dialects worldwide. Expanding the scope of our research to encompass a more extensive set of dialects would provide a more comprehensive understanding of dialect translation challenges and opportunities.
    \item Fine-Tuning Strategies: Investigating advanced fine-tuning techniques for LLMs to enhance their performance in dialect translation is an area of potential exploration. Fine-tuning models specifically for different dialects could lead to more precise and context-aware translations.
    \item User-Centric Applications: Adapting our research for practical applications is essential. Developing user-friendly tools and applications that leverage the insights from our research to provide real-time dialect translation services could bridge language gaps and improve cross-cultural communication.
    \item Human-AI Collaboration: While LLMs play a vital role in dialect translation, combining them with human linguistic expertise could be a valuable strategy. Research into the seamless collaboration of AI and human translators can further enhance translation quality.
    \item Evaluation Metrics Refinement: Developing more nuanced evaluation metrics tailored to dialect translation quality would provide a better understanding of the subtle nuances and intricacies involved. This could lead to more accurate model assessment.
\end{enumerate}}

\section{References}
\bibliographystyle{plain}
\bibliography{references}
\cite{towardsdatascience}
\cite{liu2023lost}
\cite{pile}
\cite{chen2023llama}
\cite{brown2022pathways}
\cite{baevski2021scaling}
\cite{gpt35}
\cite{chatgpt}
\cite{llama2chat}
\cite{mistralchat}
\cite{falconchat}

\end{document}